\documentclass[runningheads]{llncs}

\usepackage{eccv}

\usepackage{eccvabbrv}

\usepackage{graphicx}
\usepackage{booktabs}
\usepackage{multirow}
\usepackage{makecell}
\usepackage{bm}
\usepackage{amsmath}
\usepackage{amsfonts}
\usepackage{pifont}
\usepackage[accsupp]{axessibility}  % Improves PDF readability for those with disabilities.

\usepackage[pagebackref,breaklinks,colorlinks,citecolor=eccvblue]{hyperref}
\usepackage{orcidlink}

\begin{document}

\title{Self-learning Canonical Space for Multi-view 3D Human Pose Estimation} 


% \institute{ShanghaiTech University \and United Imaging Intelligence}
\author{Xiaoben Li\inst{1,2}\and
Mancheng Meng\inst{2} \and
Ziyan Wu\inst{2} \and Terrence Chen\inst{2} \and Fan Yang\inst{2}\thanks{Corresponding author.}\and Dinggang Shen\inst{1,2}}

\authorrunning{X.~L et al.}

\institute{ShanghaiTech University \and United Imaging Intelligence}

\maketitle

\begin{abstract}
  Multi-view 3D human pose estimation is naturally superior to single view one, benefiting from more comprehensive information provided by images of multiple views. The information includes camera poses, 2D/3D human poses, and 3D geometry. However, the accurate annotation of these information is hard to obtain, making it challenging to predict accurate 3D human pose from multi-view images. To deal with this issue, we propose a fully self-supervised framework, named cascaded multi-view aggregating network (CMANet), to construct a canonical parameter space to holistically integrate and exploit multi-view information. In our framework, the multi-view information is grouped into two categories: 1) intra-view information (\ie, camera pose, projected 2D human pose, view-dependent 3D human pose), 2) inter-view information (\ie, cross-view complement and 3D geometry constraint). Accordingly, CMANet consists of two components: intra-view module (IRV) and inter-view module (IEV). IRV is used for extracting initial camera pose and 3D human pose of each view; IEV is to fuse complementary pose information and cross-view 3D geometry for a final 3D human pose. To facilitate the aggregation of the intra- and inter-view, we define a canonical parameter space, depicted by per-view camera pose and human pose and shape parameters ($\theta$ and $\beta$) of SMPL model, and propose a two-stage learning procedure. At first stage, IRV learns to estimate camera pose and view-dependent 3D human pose supervised by confident output of an off-the-shelf 2D keypoint detector. At second stage, IRV is frozen and IEV further refines the camera pose and optimizes the 3D human pose by implicitly encoding the cross-view complement and 3D geometry constraint, achieved by jointly fitting predicted multi-view 2D keypoints. The proposed framework, modules, and learning strategy are demonstrated to be effective by comprehensive experiments and CMANet is superior to state-of-the-art methods in extensive quantitative and qualitative analysis. 
\end{abstract}

\section{Introduction}
3D human pose estimation (HPE) as a fundamental task in computer vision involves detecting the 3D locations of body keypoints and building a geometry representation of the human body from visual data \cite{Tekin2016,martinez_2017_3dbaseline, Sun2017}. It has a wide range of applications, including human-computer interaction, augmented reality, and virtual reality. Due to the critical importance and wide applications of 3D HPE, numerous works have been conducted in this field. Considerable amount of efforts are spent on recovering 3D human model (pose/shape) from a single image \cite{Chen20173DHP, Li_Chan_2014, pavlakos2017volumetric, Sun2017}. However, it is an ill-posed task because of depth ambiguity and self-occlusion. In contrast, multi-view images inherently contain more comprehensive information, so that it is more feasible to perform 3D HPE \cite{3DPictorial, Liang_2019_ICCV, Qiu_2019_ICCV}.

Although multi-view images contain crucial information for 3D HPE, such as camera pose, human pose and shape from each view, and cross-view 3D geometry, appropriately learning the information is problematic. The challenges can be concluded into two aspects: 1) the corresponding annotations of the aforementioned information are scarce and labor intensive; 2) the information is heterogeneous and hard to model in an unified space.
Consequently, we propose a cascade self-supervised framework and a canonical parameter space 
to handle the two challenges, respectively. 

\begin{figure*}[t]
  \centering
  \includegraphics[width=\linewidth]{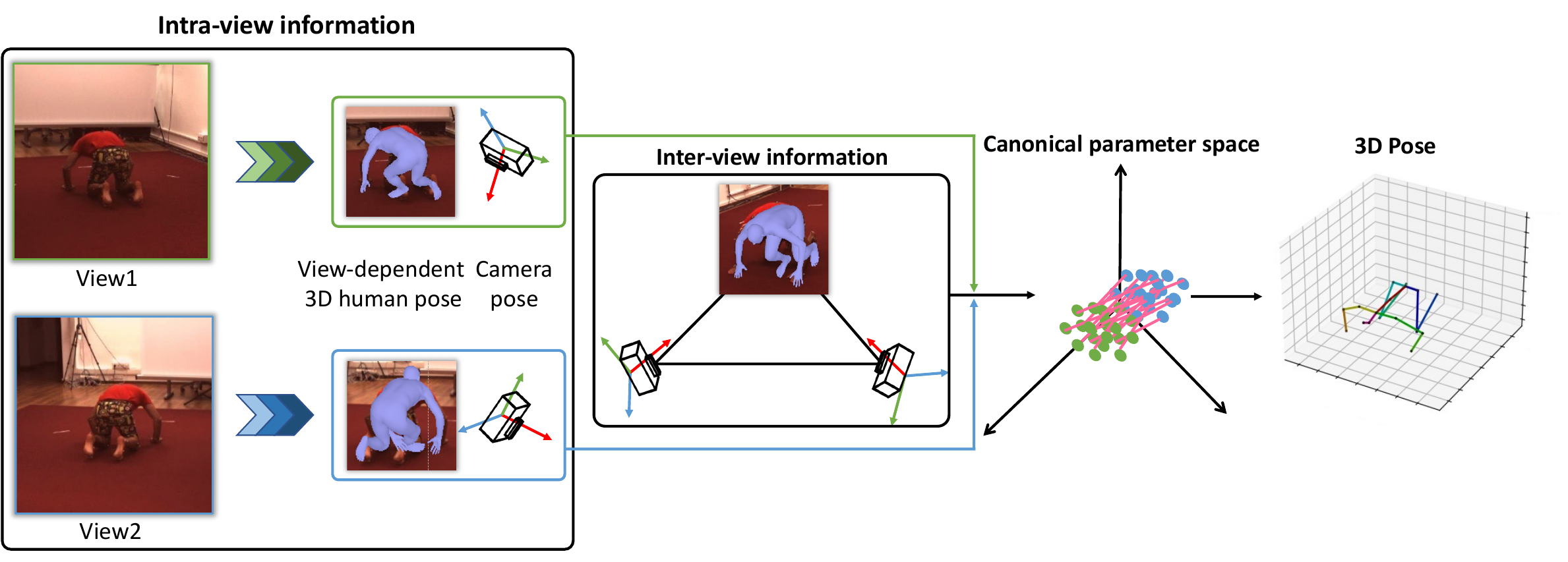}
  \caption{The illustration of the two categories of multi-view information (intra-view and inter-view) and the brief idea of proposed cascaded multi-view aggregation pipeline and canonical parameter space. For demonstration simplicity, two-view images are adopted here, in practice multi-view case is compatible.}
  \label{fig_idea}
\end{figure*}

Multi-view information is composed by multiple spatially correlated single-view images, thus in our framework it is naturally separated into intra-view (\ie, camera pose and view-dependent 3D human pose) and inter-view information (\ie, cross-view complement and 3D geometry constraint) (Figure \ref{fig_idea}). The framework, named cascaded multi-view aggregating network (CMANet), encompasses two major components: intra-view module (IRV) and inter-view module (IEV) to process the two types of information, respectively. 
Within the framework, a canonical parameter space is proposed and parameterized by camera pose and human pose and shape parameters ($\theta$ and $\beta$) of SMPL model \cite{SMPL:2015}
% \fnote{missing reference}
so that the heterogeneous information can be integrated into a homogeneous parameter space for communication (Figure \ref{fig_idea}).    

For intra-view information, we resort to 2D keypoints and SMPL model \cite{SMPL:2015} to recover the camera pose and view-dependent 3D human pose in IRV. Since the state-of-the-art 2D keypoint detectors achieve human-competitive performance, an off-the-shelf detector~\cite{sun2019deep} is adopted to yield confident 2D keypoints, screening by confidence score. Due to the differentiable parametric property of SMPL model and inspiration from HMR \cite{Kanazawa2017} and SPIN \cite{Kolotouros2019SPIN}, the camera pose and view-dependent 3D human pose can be estimated by 
minimizing reprojection error with the 2D keypoints and represented in a canonical parameter space. This procedure is employed on each view of image to obtain parameterized representation of intra-view information. 

The resulting 3D human poses are inaccurate owing to the single-view limitations: depth ambiguity and self-occlusion problems. 
Given camera pose, multi-view of image can construct the geometry of 3D space. 
The different views of image can complement each other to capture more complete human pose, especially when some keypoints are exclusively visible in certain views. Therefore, the cross-view complementary and 3D geometrical characteristics of inter-view information can serve to mitigate the limitations. To leverage the inter-view information, we aggregate the estimated  per-view camera pose and 3D human pose into IEV as the initial value of the canonical parameter space for a shallow network to refine camera poses and optimize a 3D human pose by simultaneously minimizing the reprojection error with multi-view 2D keypoints. In this way, the inter-view information is implicitly exploited in a canonical space for 3D human pose.

% The contributions of the work are summarized into three-fold: \Romannum{1}. A fully unsupervised cascaded framework together with a canonical parameter space are proposed for 3D human pose estimation; \Romannum{2}. According to the characteristics of multi-view information, a two-stage learning procedure is proposed to facilitate modeling; \Romannum{3}. Extensive experiments on large datasets validate the efficacy and superiority of the proposed components and method.

The contributions of the work are summarized into three-fold: I. A fully unsupervised cascaded framework together with a canonical parameter space are proposed for 3D human pose estimation; II. According to the characteristics of multi-view information, a two-stage learning procedure is proposed to facilitate modeling; III. Extensive experiments on large datasets validate the efficacy and superiority of the proposed components and method.
\section{Related Work}
\textbf{Single-view 3D Human Pose Estimation.} 
With the rapid development of deep neural networks, there are many works estimating 3D pose of human using single-view image. These works can be further divided into: direct estimation methods and 2D to 3D lifting methods. 
Direct estimation methods infer the 3D human pose from 2D image without intermediately estimating 2D pose representation. 
The capacity of deep neural networks for direct 3D human pose estimation is fully explored by CNN-based methods \cite{Li_Chan_2014,pavlakos2017volumetric, Sun2017} and transformer-based methods \cite{zheng20213d, zhao2023poseformerv2}. Recently, diffusion-based method is also adopted for direct 3D human pose estimation \cite{gong2023diffpose} and the field is further advancing. Although promising results have been achieved, these methods typically require 3D supervision which is not satisfied in real-world scenario.
2D to 3D lifting methods infer 3D human pose from the intermediately estimated 2D human pose \cite{Chen20173DHP, martinez_2017_3dbaseline}. This type of methods are demonstrated to be effective \cite{martinez_2017_3dbaseline} but the effect is dependent on the input 2D poses and 3D supervision is also required. Besides, 3D human mesh recovery methods \cite{Kanazawa2017, Kolotouros2019SPIN} predict human mesh that contains rich shape and pose information directly from a image. In these methods the mesh is represented as parameters of a skinned vertex-based model, SMPL \cite{SMPL:2015} .
\begin{figure*}[t]
  \centering
  \includegraphics[width=\linewidth]{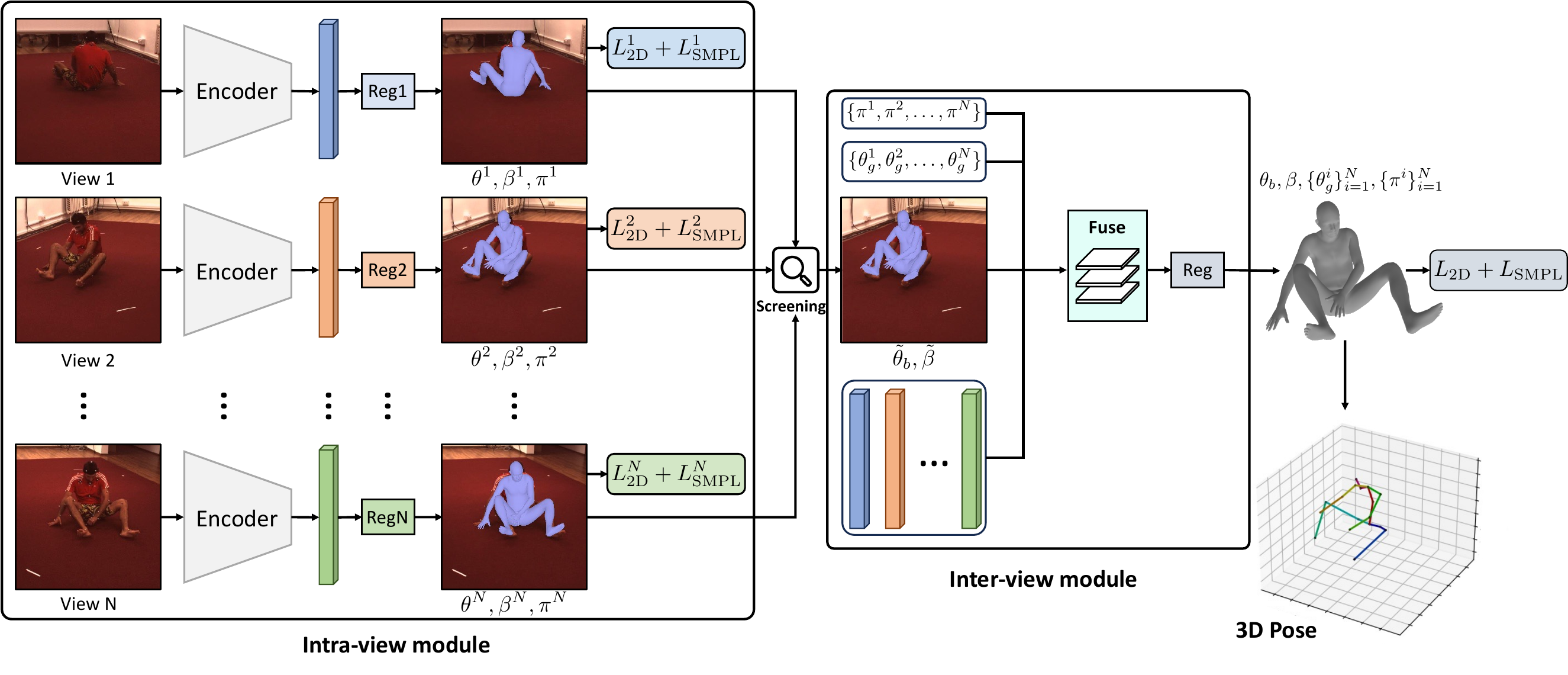}
  \caption{The architecture of the proposed cascaded multi-view aggregating network (CMANet) consisting of two components: intra-view module (IRV) and inter-view module (IEV). IRV estimates intra-view information, \ie, camera pose and 3D human pose and shape, of each view. After screening, IEV leverages the camera poses, global orientations, the optimal human body pose and shape to initialize proposed canonical parameter space, then aggregates the features from all views to refine the camera pose and human model to output final human 3D keypoints. Furthermore, a two-stage learning procedure is adopted. At the first stage IRV learns to extract intra-view information of each view, supervised by reprojection loss provided by 2D keypoints detector and SMPL loss provided by SMPLify. At the second stage IRV is frozen and IEV learns to fuse multi-view information supervised by reprojection loss and SMPL loss from all views.
  }
  \label{fig_framework}
\end{figure*}

\noindent\textbf{Multi-view 3D Human Pose Estimation.}
Although great progress have been made in single-view 3D pose estimation, these methods still suffer from occlusion and depth ambiguity. To tackle the problems, multi-view 3D pose estimation exploit geometric information from multiple views to infer 3D pose. A group of methods \cite{3DPictorial, Pavlakos_mutiview_2017_CVPR} utilize Pictorial Structure Model (PSM) to represent the human body and infer the 3D pose from 2D poses by optimizing model parameters to match the model projection with the 2D pose. However, multi-view information is not fully and effectively fused in these PSM-based methods. Therefore, many efforts have been dedicated to fuse information from multiple observations of human body \cite{Iskakov_2019_learntri, Qiu_2019_ICCV}. 
Volumetric representation \cite{Iskakov_2019_learntri} and epipolar geometry \cite{Qiu_2019_ICCV, ma2021transfusion, ma2022ppt} are explored to fuse features from different views with camera parameters and infer 3D pose through a neural network
There are also works focusing on multi-view methods without need of camera calibrations \cite{gordon2022flex, shuai2022adaptive}, which typically explore features of human body that invariant to the camera position.

\noindent\textbf{Unsupervised 3D Human Pose Estimation.}
Although deep neural networks have shown powerful capabilities in the field of 3D human pose estimation, their main drawback is the requirement of large amount of labeled data. There are large benchmarks \cite{Human3.6M, MPI-INF-3DHP} collected in constrained laboratory scenarios but few datasets are collected in unconstrained wild environment, leading to large gap between research and practice. Therefore, many efforts have attempted to explore unsupervised methods for 3D human pose estimation~\cite{Chen_2019_CVPR,Drover2018, honari2022unsupervised, Kocabas_2019_CVPR, Li_Li_Jiang_Zhang_Huang_Xu_2020,Rhodin_2018_ECCV,Rhodin_2018_CVPR}. Multi-view consistency constraint is utilized  as supervision to train a neural network \cite{Rhodin_2018_CVPR}, but the method still need a small amount of 3D ground-truth data to ensure accurate prediction. Adversarial learning strategy is also adopted to provide supervision \cite{Drover2018}. 
% To train a single-view 3D pose estimator without ground-truth 3D ?annotations, 
Multi-view information is also harvested to generate 3D annotations from multi-view input \cite{Pavlakos_mutiview_2017_CVPR} and \cite{Kocabas_2019_CVPR}.
In a recent work \cite{honari2022unsupervised}, solely multiple-view geometry is provided to estimate 3D keypoints from multi-view images. It leverages reprojection and estimation of human mask to ensure that the estimated 3D keypoints are meaningful. Creating realistic synthetic data \cite{Liang_2019_ICCV, gong2023progressive} is another way to avoid the dependence of ground-truth annotations while extra efforts are required.

\noindent\textbf{Difference from Previous Works.} 1) Our method does not need any forms of supervision from ground-truth annotations: camera pose, 2D/3D human pose. 2) We construct  a canonical parameter space to holistically integrate and exploit heterogeneous multi-view information. 3) Our method contains a two-stage learning strategy to exploit intra-view and inter-view information.

\section{Method}
\subsection{SMPL-based Human Body Representation}
Skinned Multi-Person Linear Model (SMPL)~\cite{SMPL:2015} is a skinned vertex-based model and can accurately represent a wide variety of body shapes in natural human poses. It provides a function $M(\beta, \theta;\Phi): \mathbb{R}^{|\theta|\times|\beta|}\mapsto \mathbb{R}^{3N}$, mapping shape and pose parameters to $N=6890$ vertices to form a mesh representation of human body. In the function $\beta \in \mathbb{R}^{10}$, $\theta \in \mathbb{R}^{24\times 3}$, and $\Phi$ are shape, pose and learning model parameters, respectively. The shape parameter $\beta$ is coefficient of a low-dimensional shape space, learned from a training set of thousands of registered scans. The pose parameter $\theta$ includes global rotation $\theta_g \in \mathbb{R}^{3}$ and body pose parameter $\theta_b \in \mathbb{R}^{23\times 3}$, consequently $\theta = \{\theta_g, \theta_b\}$.

Typically, a pre-trained linear regressor $W$ is used to obtain $k$ keypoints of interest of the whole body, \ie, $J_{3D}\in \mathbb{R}^{k\times 3} = WM(\beta, \theta;\Phi)$ \cite{Kolotouros2019SPIN}. As for 2D keypoints, a perspective camera model with fixed focal length and intrinsics $K$ is used to project 3D joint positions into the 2D image plane. Each camera pose $\pi = \{R, t\}$ consists of global orientation $R \in \mathbb{R}^{3}$ and translation $t\in\mathbb{R}^3$.
Given above parameters, 3D keypoints can be projected to the 2D image as  $J_{2D} = \Pi (K(RJ_{3D}+t))$, where $\Pi$ is a perspective projection with camera intrinsics $K$. Since the pose parameter $\theta$ already includes a global orientation $\theta_g$, in practice we assume $R$ of the camera model as identity matrix and only predict camera translation $t$. 

\subsection{Cascaded Multi-view Aggregating Network}

The architecture of the proposed cascaded multi-view aggregating network (CMANet) is depicted in Figure \ref{fig_framework}, consisting of two major components: intra-view module (IRV) and inter-view module (IEV). Formally, given $N$-view images $\{I^i\}_{i=1}^N$ of a person, IRV firstly extracts camera pose, 3D human model and image features of each view $\{\theta^i, \beta^i, \pi^i, F^i\}_{i=1}^N$ with $N$ shared encoders and $N$ regressors. IEV leverages the camera poses $\{\pi^i\}_{i=1}^N$, global orientations $\{\theta_g^i\}_{i=1}^N$, the optimal human body pose and shape $\tilde{\theta}_{b}$ and $\tilde{\beta}$ screened from all views as initial parameter of the canonical space, and aggregates the features of each view into a light-weight sub-network to further optimize human model and refine camera poses. In learning phase, the weights of IRV is first optimized independently, then it is frozen and IEV learns to estimate final human pose. In addition, all $\beta, \theta, \pi$ are supervised by deteteced 2D keypoints.
\subsection{Intra-view Module}
This module aims to extract intra-view information, \ie, camera pose ($\pi = \{ R, t \}$) and 3D human pose and shape ($\{\theta, \beta\}$) of each view. Since intra-view information is contained in each single view of image and independent from cross-view geometry, each view of image is processed independently. More specifically,  given $N$ views of images $\{I^i\}_{i=1}^N$ of a human body, the images are first forwarded into $N$ weights-shared encoders for feature extraction. Subsequently, the extracted feature map $F^i$ of $i$-th image is reshaped into a fixed-length feature vector and fed into a corresponding $i$-th regressor to yield $i$-th camera pose $\pi^i$, 3D human pose $\theta^i$ and shape $\beta^i$ of $i$-th view.

Due to the powerful feature representation capacity of Swin-Transformer~\cite{Liu_2021_ICCV_swin}, we adopt it as architecture of the encoder. Considering the success of previous work~\cite{Kanazawa2017}, instead of directly regressing parameters in one go, we regress camera pose and human model parameters in an iterative manner, where mean value of camera pose and human pose and shape parameters are set as the initial value of parameters. We further utilize the representation proposed by~\cite{Zhou_2019_CVPR} for the 3D rotations as in~\cite{Kolotouros2019SPIN} for faster convergence during learning. The $N$ regressors are constructed by $N$ weights-unshared two-layer fully connected network.

\subsection{Inter-view Module}
 The module is to extract inter-view information, \ie, cross-view complement and 3D geometry constrain, and fuse complementary intra-view human and camera pose information and cross-view 3D geometry. To facilitate the aggregation of the intra- and inter-view information, we define a holistic canonical parameter space for multi-view pose representation, which contains per-view camera pose $\{\pi^i\}_{i=1}^N=\{ R^i, t^i \}_{i=1}^N$, global orientations of each view $\{\theta_g^i\}_{i=1}^N$, and view-independent shape and body pose parameters $\{\theta_b, \beta \}$. Note that in multi-view scenarios, different views correspond to the same human body thus have the same shape and body pose parameters $\{\theta_b, \beta\}$, which are view-independent. On the contrary, the camera poses of different views differ from each other, so we have $N$ global orientations $\{\theta_g^i\}_{i=1}^N$ ($R_i$ is assumed to be identity matrix).
 \begin{table}[t]
\centering
\renewcommand{\arraystretch}{1.2}
\caption{Ablation study of each component in our CMANet on MPI-INF-3DHP dataset.}
% \resizebox{\linewidth}{!}{ 
\begin{tabular}{ccccccccc}
\toprule
\multicolumn{3}{c}{Method}& \multicolumn{6}{c}{Metrics}                                        \\ \cline{1-9} 
             \multirow{2}{*}{IEV} & \multirow{2}{*}{IRV} & \multirow{2}{*}{\makecell{Two-stage\\ learning}}           & \multicolumn{3}{c}{Absolute} & \multicolumn{3}{c}{Rigid Alignment} \\ \cline{4-9} 
                          &  & 
                         & PCK $\uparrow$    & AUC $\uparrow$    & MPJPE $\downarrow$   & PCK $\uparrow$      & AUC $\uparrow$      & MPJPE   $\downarrow$    \\
                         \midrule
                 {\color{red}\ding{51}}   & {\color{green}\ding{55}} & {\color{green}\ding{55}}
                      &     82.0   &     30.7    &     117.8    &   87.1        &        51.6   &       75.8     \\
                     {\color{red}\ding{51}}  & {\color{red}\ding{51}} & {\color{green}\ding{55}}
                        &   79.1     &    29.3     &     135    &  87.0         &      51.2    &     79.8      \\
                      {\color{red}\ding{51}}   &{\color{red}\ding{51}}  &{\color{red}\ding{51}} 
                        &      \textbf{89.1}   &    \textbf{46.1}     &     \textbf{93.7}     &        \textbf{93.9}   &      \textbf{59.2}     &  \textbf{68.2} \\
                         \bottomrule
\end{tabular}
% }

\label{ablation_component_mpii3d}
\end{table}
\begin{table}[t]
\centering
\renewcommand{\arraystretch}{1.2}
\caption{Ablation study of visible keypoints for learning on MPI-INF-3DHP dataset. `Min\_Max' indicates that we use the view that have most visible keypoints and the view that have fewest visible keypoints, and `Min\_2' indicates that we use the 2 views that have fewest visible keypoints.}
% \resizebox{\linewidth}{!}{ 
\begin{tabular}{ccccccc}
\toprule
\multirow{3}{*}{Num of views} & \multicolumn{6}{c}{Metrics}                                        \\ \cline{2-7} 
                         & \multicolumn{3}{c}{Absolute} & \multicolumn{3}{c}{Rigid Alignment} \\ \cline{2-7} 
                         & PCK $\uparrow$    & AUC $\uparrow$    & MPJPE  $\downarrow$  & PCK   $\uparrow$    & AUC  $\uparrow$     & MPJPE  $\downarrow$     \\
                         \midrule
             Min\_Max         &     70.1    &    27.8     &      154.4    &      85.0    &    42.3       &     100.6        \\
             % best 122.2936947722557 90.90367821251734
                Min\_2       &    65.4     &    27.2     &     168.5     &       79.4    &     38.0      &    107.3         \\
                % best 117.61903729237314 97.80315685786802 
                      4   &      \textbf{89.1}   &    \textbf{46.1}     &     \textbf{93.7}     &        \textbf{93.9}   &      \textbf{59.2}     &  \textbf{68.2}     \\
                         \bottomrule
\end{tabular}
% }

\label{tab:ablation_mpii3d}
\end{table}
\begin{table*}[t]
\centering
\renewcommand{\arraystretch}{1.2}
\caption{Comparisons on MPJPE and PA-MPJPE (both in mm) on the Human3.6M test set with other methods. `Multi-view' indicates methods which take multi-view images or videos as input. `Cameras' indicates methods which need camera extrinsic parameters. `Temporal' indicates methods which use temporal information, \ie, videos for human pose/shape estimation.
}
\resizebox{\linewidth}{!}{ 
\begin{tabular}{ccccccc}
\toprule
\multirow{2}{*}{Method} & \multirow{2}{*}{Supervision} & \multirow{2}{*}{Multi-view} & \multirow{2}{*}{Cameras}  &  \multirow{2}{*}{Temporal} & \multicolumn{2}{c}{Metrics} \\ \cline{6-7} 
                         &                             &            
                         &                                      &            & MPJPE $\downarrow$      & PA-MPJPE $\downarrow$     \\ \hline
             Rhodin \textit{et al.} \cite{Rhodin_2018_ECCV}                &       J3D                      &         {\color{red}\ding{51}}                    &              {\color{red}\ding{51}}            &                       {\color{green}\ding{55}}      &    131.70   &       98.20        \\
         HMR \cite{Kanazawa2017}               &        J2D/Adv.        &    {\color{green}\ding{55}}         &          {\color{green}\ding{55}}                    &        {\color{green}\ding{55}}              &           106.84           &          67.45     \\ 
SPIN \cite{Kolotouros2019SPIN} &                       J2D      &     {\color{green}\ding{55}}       &     {\color{green}\ding{55}}             &                          {\color{green}\ding{55}}          &                  -      &       62.00        \\ 
Liang \textit{et al.} \cite{Liang_2019_ICCV}                &           J2D/J3D/Mesh                  &                   {\color{red}\ding{51}}           &     {\color{green}\ding{55}}                     &                  {\color{green}\ding{55}}                   &    79.85     &       45.13        \\
Qiu \textit{et al.} \cite{Qiu_2019_ICCV}&                        J2D     &      {\color{red}\ding{51}}      &   {\color{red}\ding{51}}               &                            {\color{green}\ding{55}}            &                     26.20      &      -         \\ 
VIBE \cite{kocabas2020vibe}&                        J2D/J3D/Mesh/Adv.     &       {\color{green}\ding{55}}     &     {\color{green}\ding{55}}             &                        {\color{red}\ding{51}}         &        65.60       &          41.40     \\ 
AdaFuse \cite{zhang2020adafuse}&                        J2D     &      {\color{red}\ding{51}}      &   {\color{red}\ding{51}}               &                            {\color{green}\ding{55}}            &                 19.50      &      -         \\ 
Shin \textit{et al.} \cite{shin2020multiview}&                  J2D/J3D/Mesh           &      {\color{red}\ding{51}}      &      {\color{red}\ding{51}}            &                                 {\color{green}\ding{55}}            &                      46.90     &      32.50        \\ 
  Li \textit{et al.} \cite{Li_2021_WACV} &                      J2D/J3D      &     {\color{red}\ding{51}}                         &         {\color{green}\ding{55}}                 &                 {\color{green}\ding{55}}      &             64.80     &   43.80            \\ 
  ProHMR \cite{kolotouros2021prohmr}                &           J2D/J3D/Mesh                 &    {\color{green}\ding{55}}                          &    {\color{green}\ding{55}}                      &            {\color{green}\ding{55}}             &    62.20   &       34.50        \\
          TesseTrack \cite{reddy2021tessetrack}                &            Mesh                 &               {\color{red}\ding{51}}               &          {\color{red}\ding{51}}                &       {\color{red}\ding{51}}                  &     18.70   &          -     \\
         Shuai \textit{et al.}\cite{shuai2022adaptive}                &          J3D                   &        {\color{red}\ding{51}}                      &                {\color{green}\ding{55}}          &        {\color{red}\ding{51}}                       &    26.20   &       -        \\
          Chun \textit{et al.}\cite{Chun_2023_WACV}&              Mesh    &      {\color{red}\ding{51}}     &    {\color{red}\ding{51}}                          &       {\color{green}\ding{55}}             &                17.59   &      -         \\ 
         \midrule
         Kocabas \textit{et al.} \cite{Kocabas_2019_CVPR}                         &      Self         &  {\color{red}\ding{51}}            &      {\color{green}\ding{55}}                        &          {\color{green}\ding{55}}              &                      76.60       &       67.45        \\ 
         Kundu \textit{et al.} \cite{Kundu_2020_CVPR}                         &       Self        &   {\color{green}\ding{55}}           &        {\color{green}\ding{55}}                      &        {\color{green}\ding{55}}           &                           99.20      &     -          \\
         Kundu \textit{et al.} \cite{kundu2020kinematic}                         &      Self         &  {\color{green}\ding{55}}            &        {\color{green}\ding{55}}                      &    {\color{green}\ding{55}}             &                        -     &       89.40        \\
         Honari \textit{et al.} \cite{honari2022unsupervised}                &             Self                &             {\color{red}\ding{51}}                 &                {\color{red}\ding{51}}          &                          {\color{green}\ding{55}}       &     73.80    &    63.00           \\ 
          Honari \textit{et al.} \cite{Honari_2023_Temporal}                &         Self                    &                {\color{green}\ding{55}}              &          {\color{green}\ding{55}}                &                   {\color{red}\ding{51}}        &     100.30    &    74.90           \\
           Ours            &          Self                   &         {\color{red}\ding{51}}                     &        {\color{green}\ding{55}}                   &        {\color{green}\ding{55}}         &       \textbf{64.48}          &    \textbf{51.50}           \\ \bottomrule
\end{tabular}
}

\label{compare_h36m}
\end{table*}
\begin{table*}[t]
\centering
\caption{Comparison results on TotalCapture test set in terms of 3D pose errors MPJPE (mm). Note that IMU-PVH \cite{TotalCapture} uses information from the IMU sensors as supervision.}
\resizebox{\linewidth}{!}{ 
\begin{tabular}{ccccccccc}
\toprule
\multirow{2}{*}{Method} & \multirow{2}{*}{Supervision} & \multicolumn{3}{c}{Subjects 1,2,3} & \multicolumn{3}{c}{Subjects 4,5} &  \multirow{2}{*}{Mean}\\
                  &                   &    Walking 2   &   Acting 3    &  Freestyle 3    &  Walking 2   &   Acting 3    &  Freestyle 3    &  \\ \midrule
                 Tri-CPM \cite{wei2016convolutional}&    3D               &  79.0     &    106.5   &   112.1   &   79.0    &   73.7    &   149.3   &  99.8\\
                 PVH \cite{TotalCapture}&         3D          &    48.3   & 94.3      &   122.3   &  84.3     &  154.5     &    168.5  & 107.3 \\
                  IMU-PVH \cite{TotalCapture}&             3D      &   30.0    &    49.0   &  90.6    &   36.0    &     109.2  &   112.1   & 70.9 \\
                 Trumble \textit{et al.} \cite{trumble2018deep} &  3D  &     42.0          &   59.8    &   120.5    & 58.4     &  103.4     &   162.1    &   85.4  \\
                 ProHMR \cite{kolotouros2021prohmr}  &   Mesh                &   125.7    &   118.9    &   134.3   &   131.9    &   125.2    &    135.8  & 127.8 \\
            Ours      &      Self             &   \textbf{76.6}    &  \textbf{74.2}     &   \textbf{74.3}   &   \textbf{80.8}    &   \textbf{77.2}    &   \textbf{86.7}   & \textbf{77.9} \\ \bottomrule
\end{tabular}
    }

\label{compare_total_capture}
\end{table*}
\begin{table}[t]
\centering
\renewcommand{\arraystretch}{1.2}
% \resizebox{\linewidth}{!}{ 
\caption{Comparison results on MPI-INF-3DHP with and without rigid alignment. The top half is methods using different forms of supervision while the bottom half is self-supervised methods.}
\begin{tabular}{ccccccc}
\toprule
\multirow{3}{*}{Method} & \multicolumn{6}{c}{Metrics}                                        \\ \cline{2-7} 
                         & \multicolumn{3}{c}{Absolute} & \multicolumn{3}{c}{Rigid Alignment} \\ \cline{2-7} 
                         & PCK $\uparrow$    & AUC$\uparrow$      & MPJPE$\downarrow$    & PCK $\uparrow$       & AUC $\uparrow$       & MPJPE$\downarrow$     \\
                         \midrule
                   HMR \cite{Kanazawa2017}        &     59.6    &      27.9   &        169.5  &     77.1      &       40.7    &   113.2          \\
                  SPIN \cite{Kolotouros2019SPIN} &   66.8    &          30.2 &   124.8      &    87.0      &    48.5              &       80.4      \\
                  Liang \textit{et al.} \cite{Liang_2019_ICCV}   &   -      &     -    &    -      &      95     &     63      &      62      \\
                   VIBE \cite{kocabas2020vibe}  &    -     &     -    &      96.6    &   89.3        &    -       &        64.6     \\
                 Shin \textit{et al.} \cite{shin2020multiview}  &   -      &     -    &    -      &      97.4     &     65.5      &      50.2       \\
                  Li \textit{et al.} \cite{Li_2021_WACV}   &  79.2       &    39.3     &    98.7      &     92.9      &   56.1        &     65.6        \\
                 ProHMR \cite{kolotouros2021prohmr}  &   -      &      -   &     -     &     -      &   -        &       65.0      \\
                 Chun \textit{et al.} \cite{Chun_2023_WACV}   &    94.1     &    59.3    &     68.0    &     -      &       -    &        -     \\ \midrule
                  Kocabas \textit{et al.} \cite{Kocabas_2019_CVPR}     &    64.7     &    -     &    126.79      &     -      &     -      &       -      \\
                  Kundu \textit{et al.} \cite{Kundu_2020_CVPR}      &      -   &   -      &  -        &    83.2       &58.7&         97.6    \\
                   Kundu \textit{et al.} \cite{kundu2020kinematic}       &   -      &   -      &     -     &      79.2     &     43.4      &     99.2        \\
                Honari \textit{et al.} \cite{honari2022unsupervised}        &     -    &     -    &     145.7     &    -       &        -   &       115.2      \\
                   Honari \textit{et al.} \cite{Honari_2023_Temporal}      &   -     &    -     &     209.5    &   -        &      -     &         140.4   \\
                      Ours   &      \textbf{89.1}   &    \textbf{46.1}     &     \textbf{93.7}     &        \textbf{93.9}   &      \textbf{59.2}     &  \textbf{68.2} \\
                         \bottomrule
\end{tabular}
% }

\label{compare_mpii3d}
\end{table}

As shown in Figure \ref{fig_framework}, after IRV extracts camera pose and human pose of each view, we forwarded the optimal human pose, shape, all camera poses and global orientations into aforementioned canonical parameter space.
The screened parameters together with features from all views are fused by a single Swin Transformer layer (STL), following by a regressor to output the final human pose, shape and camera poses.
Specifically, the IRV estimated camera pose parameters $\{ R^i, t^i \}_{i=1}^N$ and global orientations $\{\theta_g^i\}_{i=1}^N$ are used as the initial for regression in IEV. From $\{\theta^i_b, \beta^i\}_{i=1}^N$, we select the one with least 2D reprojection error as the initial view-independent shape and body pose parameters $\{ \tilde{\theta_{b}}, \tilde{\beta} \}$. The features $\{F^i\}_{i=1}^N$ from all views are fused by a Swin Transformer layer and forwarded into a regressor to refine human model and camera pose, so as to output final human pose.

\subsection{Self-supervised Two-stage Learning}
In our CMANet, intra-view module (IRV) and intra-view module (IEV) exploit intra-view and inter-view information, respectively. As the two modules are designed cascadedly and decoupledly, it is natural to optimize the two module in a staged manner. Specifically, we propose a two-stage learning procedure. At the first stage, IRV learns to estimate camera pose and view-dependent 3D human pose supervised by confident output of an off-the-shelf 2D keypoint detector. At the second stage, IRV is frozen and IEV further learns to refine the camera pose and optimize the 3D human pose by implicitly encoding the cross-view complement and 3D geometry constraint, achieved by jointly fitting the estimated multi-view 2D keypoints.

In learning phase, no ground truth 2D or 3D annotations are involved, instead, we adopt a strong 2D keypoint detector~\cite{sun2019deep} to yield body keypoints $\hat{\bm{J}}_{\rm 2D}=\{\hat{J}_{\rm 2D}^i\}_{i=1}^N$ as well as corresponding confidence scores $\bm{C}=\{C^i\}_{i=1}^N$. Then, we adopt following strategies to process the keypoint information. First, we filter out the keypoints whose confidence scores are less than a certain threshold $\lambda$ and obtain a new set of confidence scores $\bm{C}_{\rm Filt}=\{C_{\rm Filt}^i\}_{i=1}^N$.
Second, when computing 2D reprojection loss, we use the confidence scores to reweight keypoint-wise loss, \ie, the keypoints with higher confidence score have more contribution to total loss. 
At the first stage, IRV is optimized using 2D reprojection loss function. Given predicted camera pose $\{ R^i, t^i \}$ and SMPL parameters $\beta^i, \theta^i$ for $i$-th view, 2D keypoints of the views $J_{\rm 2D}^i$ are calculated by regression and projection, the 2D reprojection loss function is defined as
\begin{equation}
    % L_{\rm 2D}= \sum_i^N C_{\rm Filt}^i\| \hat{J}_{\rm 2D}^i-J_{\rm 2D}^i\|^2_2
    L_{\rm 2D}^i=  C_{\rm Filt}^i\| \hat{J}_{\rm 2D}^i-J_{\rm 2D}^i\|^2_2
    \label{2D_loss}
\end{equation}
However, as argued by~\cite{Kolotouros2019SPIN}, this 2D reprojection supervisory signal is not strong enough for the network to learn. Thus we introduce a SMPLify~\cite{Bogo:ECCV:2016} optimization routine in the learning loop. 
The loss function for SMPL parameter is defined as
\begin{equation}
    % L_{\rm SMPL}^1= \sum_i^N \|\{ \beta_{\rm opt}^i, \theta_{\rm opt}^i\}-\{ \beta_i, \theta_i\} \|^2_2
    L_{\rm SMPL}^i=\|\{ \beta_{\rm opt}^i, \theta_{\rm opt}^i\}-\{ \beta^i, \theta^i\} \|^2_2
\end{equation}
The total loss function for $i$-th view at the first stage is
\begin{equation}
    \begin{aligned}
        L_1^i = \alpha L_{\rm 2D}^i + \gamma L_{\rm SMPL}^i
    \end{aligned}
    \label{eq:loss_1}
\end{equation}
where $\alpha$ and $\gamma$ are hyper-parameters to balance the 2D reprojection loss and SMPL parameter loss. Note that at the first stage, we have $N$ loss functions for $N$ views, each of them is used to supervise the corresponding regressor and shared encoder.

At the second learning stage, the weights of IRV is frozen, and we optimize IEV using 2D reprojection loss from $N$ views.
% \fnote{In my understanding, the 2D reprojecttion at second stage is different from that at first stage.} 
The difference between the 2D reprojection loss used here and that at the first stage is that instead of backpropagating loss for each view separately, we update weights of IEV using summed 2D reprojection loss from all views. In this way, the Swin Transformer layer and regressor can implicitly model the complementary human pose and the 3D geometry constraint. Therefore, the 2D reprojection loss function for the second learning stage is defined as:
\begin{equation}
    L_{\rm 2D}= \sum_i^N C_{\rm Filt}^i\| \hat{J}_{\rm 2D}^i-J_{\rm 2D}^i\|^2_2
\end{equation}

To utilize additional prior information as supervision, we also introduce a optimization routine at the second stage, which is a multi-view adaptation of SMPLify \cite{Bogo:ECCV:2016}, similar to \cite{Li_2021_WACV}. The SMPL parameter loss function for the second stage is defined as
\begin{equation}
\begin{aligned}
    L_{\rm SMPL}=&\|\{ \beta_{\rm opt}, (\theta_b)_{\rm opt}\}-\{ \beta, \theta_b\} \|^2_2+\\
    &\sum_i^N\|(\theta^i_g)_{\rm opt}-\theta^i_g\|^2_2
    \end{aligned}
\end{equation}
The total loss function for the second stage is:
\begin{equation}
    \begin{aligned}
        L_2 = \alpha L_{\rm 2D} + \gamma L_{\rm SMPL}
    \end{aligned}
\end{equation}
where $\alpha$ and $\gamma$ are identical to the ones in the loss function for the first stage.

\section{Experiments}
\subsection{Experimental Setup}
\noindent\textbf{Datasets.}
We evaluate our method on Human3.6M~\cite{Human3.6M}, MPI-INF-3DHP~\cite{MPI-INF-3DHP} and TotalCapture~\cite{TotalCapture} datasets. Human3.6M is a large 3D human pose benchmark which contains 11 different subjects and each subject performs 15 different daily actions indoors. The subjects are captured by 4 synchronized 50 Hz digital cameras. The dataset provides 2D/3D keypoint annotations obtained by marked-based motion capture system. 
% Similar to previous work \cite{Kanazawa2017,honari2022unsupervised}, we follow the protocol 1 to use subjects 1, 5, 6, 7, and 8 for training and subjects 9 and 11 are used for validation.
We use subjects 9 and 11 in protocol 1 for model learning and validation.
MPI-INF-3DHP contains 8 actors (4 males and 4 females) performing 8 human activities. 2D and 3D keypoint annotations are also provided by marker-less motion capture system. The dataset contains multi-view images of 14 cameras with regular lenses, and we choose views 0, 2, 7, 8 from them. In addition, the standard test dataset of MPI-INF-3DHP only have single-view images, so we use 8 for learning and validation, similar to~\cite{shin2020multiview,Chun_2023_WACV}. TotalCapture dataset consists of 1.9 million frames captured from 8 calibrated full HD video cameras recording at 60Hz. The images of “Freestyle3”, “Walking2” and “Acting3” on subjects 1,2,3,4 and 5 are used for our experiments. In addition, these images from four cameras (1,3,5,7) in experiments and are without the IMU sensors.
% Following the previous work \cite{TotalCapture}, the training set consists of “ROM1,2,3”, “Walking1,3”, “Freestyle1,2”, “Acting1,2”, “Running1” on subjects 1,2 and 3. The testing set consists of “Freestyle3”, “Walking2” and “Acting3” on subjects 1,2,3,4 and 5. We use the images from four cameras (1,3,5,7) in experiments and do not use the IMU sensors.

\noindent\textbf{Evaluation Metrics.} For Human3.6M dataset, we use mean per joint position error (MPJPE) and reconstruction error as metrics. Reconstruction error is MPJPE after rigid alignment of the prediction with ground truth via Procrustes. The reconstrution error is also known as PA (procrustes aligned)-MPJPE. For MPI-INF-3DHP, this dataset is collected indoors and outdoors with a multi-camera marker-less MoCap system. Thus, the ground truth 3D annotations have some noise. In addition to MPJPE, the Percentage of Correct Keypoints (PCK) thresholded at 150mm and the Area Under the Curve (AUC) over a range of PCK thresholds before and after rigid alignment are both reported. For TotalCapture dataset, we report mean per joint position error (MPJPE) of different subjects and different action sequences as previous work~\cite{TotalCapture}.

\noindent\textbf{Implementation Details.}
In our implementation, input size of all images is 256$\times$256. IRV module is optimized for 10 epochs at the first stage and IEV is optimized for 20 epochs at the second stage with a fixed learning rate $1e-5$. All the experiments are performed on 4 NVIDIA Quadro RTX 6000 GPUs using PyTorch.
\subsection{Ablation Study}
\noindent\textbf{The Effect of Each Component.} We perform ablation study on MPI-INF-3DHP dataset. First, in Table~\ref{ablation_component_mpii3d} we evaluate each component of CMANet, including an intra-view module (IRV), an inter-view module (IEV) and two-stage learning procedure. Observing the 1st row of \ref{ablation_component_mpii3d}, IEV alone is not enough to produce accurate 3D pose. However, when adding IRV without using two-stage learning, the performance decrease notably, we attribute the reason to follows. Although IRV can aggregate information of multi-view, but it is harder to optimize IEV and IRV simultaneously as cross-view complement and 3D geometry constraint is not easy to directly explored by neural network. When introducing the proposed two-stage learning strategy, we observe apparent improvement in performance, which demonstrates the correctness of aforementioned reasoning.

\noindent\textbf{The Effect of Visible Keypoints.} We explore how the number of visible views of keypoint affect the accuracy of 3D pose in Table~\ref{tab:ablation_mpii3d}. We call one keypoint `visible' in a certain view if its confidence value is higher than a threshold, that means if a keypoint is visible in a certain view, it can provide 2D reprojection supervision. We design two different settings: `Min\_Max' and `Min\_2'. `Min\_Max' indicates that we use the view that have most visible keypoints and the view that have fewest visible keypoints, and `Min\_2' indicates that we use the 2 views that have fewest visible keypoints. The results illustrate that more visible keypoints used for learning is essential to obtain promising results. In addition, even in fewest visible scenarios, our method can still generate reasonable results, which can be attributed to the effective fusion of complementary cross-view information in the proposed canonical space.
% and the `quantity', \ie, the number of visible keypoints of the view matters.

\subsection{Comparison with State-of-the-art Methods}
\textbf{Human3.6M.} In Table \ref{compare_h36m}, we compare our method with state-of-the-art methods on test set of Human3.6M Protocol 1. Because Human3.6M is such a popular public benchmark, many works report results on it. So we distinguish these methods in different ways, \eg, whether using multi-view information, the type of supervision. Compared to other self-supervision methods, our method not only is superior, but also recovers human mesh beyond human pose while not requiring any additional information. In addition to this, our method is comparable to some supervised human mesh recovery methods like~\cite{Li_2021_WACV, kolotouros2021prohmr}. 

\noindent\textbf{MPI-INF-3DHP.} Comparison with state-of-the-art methods on MPI-INF-3DHP dataset is listed in Table~\ref{compare_mpii3d}. Similar to the result on Human3.6M, our method outperforms other self-supervised methods and is comparable to some of supervised methods (including mesh recovery methods) on all metrics (PCK, AUC and MPJPE).
\begin{figure}[t]
  \centering
  \includegraphics[width=\linewidth]{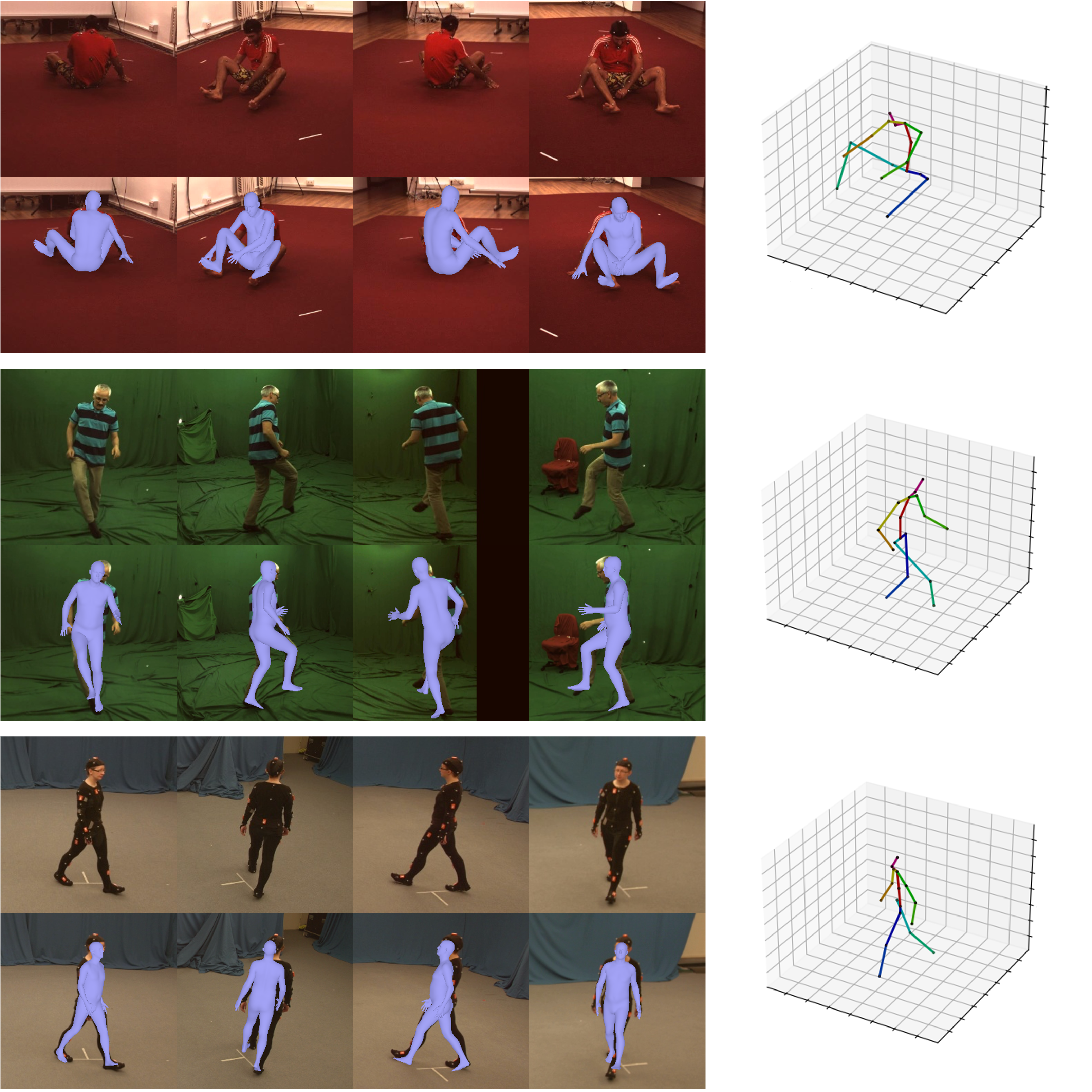}
  \caption{Visualization of reconstructed human mesh and 3D pose of samples from Human3.6M (row 1), MPI-INF-3DHP (row 2) and TotalCapture (row 3).}
  \label{qualitative results}
\end{figure}

\noindent\textbf{TotalCapture.} The comparison results with state-of-the-art methods on TotalCapture dataset are listed in Table \ref{compare_total_capture}. All methods excepts ours use various forms of supervision and IMU-PVH \cite{TotalCapture} uses
information from the IMU sensors. Our method outperforms other supervised methods that do not use IMU sensors and is comparable to IMU-PVH \cite{TotalCapture}.

\subsection{Qualitative Results}
Figure \ref{qualitative results} presents qualitative results of examples from the three datasets used in experiments. We show the reconstructed human mesh and the final 3D pose. Notice that the results are promising both for mesh and 3D pose even in the presence of self-occlusion, demonstrating the effectiveness of our method to a certain extent.

\section{Conclusion}
This paper presents a novel fully self-supervised framework, cascaded multi-view aggregating network (CMANet), for multi-view human pose estimation, where a canonical parameter space is constructed to learn multi-view information holistically. CMANet includes an intra-view module to extract view-dependent information from each view and an inter-view module to fuse multiple intra-view information, which are complementary to each other. In addition, a two-stage learning strategy is designed to facilitate CMANet to model human pose more effectively. 
% It is worth mentioning that neither any 2D/3D or mesh ground-truth annotations nor extra synthetic data is needed in CMANet, making it a large application potential. 
In future work, more prior information and better fusion strategy could be explored to further improve the performance of pose estimation.
% \bibliography{ref}
\bibliographystyle{splncs04}
\bibliography{main}

\end{document}